# Deep learning approaches in food recognition

Chairi Kiourt, George Pavlidis and Stella Markantonatou

Athena Research Centre, University Campus at Kimmeria, Xanthi, GR-67100, Greece
{chairiq, gpavlid, marks}@athenarc.gr

**Abstract**. Automatic image-based food recognition is a particularly challenging task. Traditional image analysis approaches have achieved low classification accuracy in the past, whereas deep learning approaches enabled the identification of food types and their ingredients. The contents of food dishes are typically deformable objects, usually including complex semantics, which makes the task of defining their structure very difficult. Deep learning methods have already shown very promising results in such challenges, so this chapter focuses on the presentation of some popular approaches and techniques applied in image-based food recognition. The three main lines of solutions, namely the design from scratch, the transfer learning and the platform-based approaches, are outlined, particularly for the task at hand, and are tested and compared to reveal the inherent strengths and weaknesses. The chapter is complemented with basic background material, a section devoted to the relevant datasets that are crucial in light of the empirical approaches adopted, and some concluding remarks that underline the future directions.

Keywords: Deep learning, Food recognition, Food image datasets, Convolutional neural networks.

## 1. Introduction

The advent of deep learning technology has boosted many scientific domains by providing advanced methods and techniques for better object prediction and recognition, based on images or video. In computer science, especially in computer vision and artificial intelligence, image classification is an essential task, with many recent advances coming from object recognition with deep learning approaches (Krizhevsky, Sutskever, & Hinton, 2012; Simonyan & Zisserman, 2015; He, Zhang, Ren, & Sun, 2016; Szegedy, et al., 2015; Szegedy, Vanhoucke, Ioffe, Shlens, & Wojna, 2016; Szegedy, Ioffe, Vanhoucke, & Alemi, 2017).

Food, which is an important part of everyday life in all cultures, constitutes a particular challenge in the domain of image classification, due to its visually intricate complexity, as well as the semantic complexity stemming from the variation in the mixing of various ingredients practiced by regional communities. This challenge has been proved to have many complicated aspects (Weiqing, Shuqiang, Linhu, Yong, & Ramesh, 2019; Mezgec & Koroušić, 2017). The abundance of food images provided by the social networks, dedicated photo sharing sites, mobile applications and powerful search engines (Mezgec & Koroušić, 2017) is considered to be an easy way for the development of new datasets for scientific purposes. The scientific community suggests that automatic recognition (classification) of dishes would not only help people effortlessly organize their enormous photo collections and help online photo repositories make their content more accessible, but it would also help to estimate and track daily food habits and calorie intake and plan nutritional strategies even outside a constrained clinical environment (Mezgec & Koroušić, 2017).

Despite the dozens of applications, algorithms and systems available, the problem of recognizing dishes (food) and their ingredients has not been fully addressed by the machine learning and computer vision communities (Weiqing, Shuqiang, Linhu, Yong, & Ramesh, 2019). This is due to the lack of distinctive and spatial layout of food images, typically found in images depicting scenes or objects. For example, an image of an outdoor scene can be typically decomposed into (a) a ground place, (b) a horizon, (c) a forest, and (d) the sky. Patterns like these, cannot be found in food images. Food ingredients, like those found in a salad, are mixtures that frequently come in different shapes and sizes, depending much on regional practices and cultural habits. It should be highlighted that the nature of the dishes is often defined by the different colors, shapes and textures of the various ingredients (Ciocca, Napoletano, & Schettini, 2018). Nevertheless, the kind of features that describe food in most cases are easily recognizable by humans in a single image, regardless of the geometrical variations of the ingredients. Hence, it could be considered that food recognition is a specific and a very complex classification problem demanding the development of models that are able to exploit local information (features) within images, along with complex, higher level semantics. Food recognition is still a challenging problem that attracts the interest of the scientific community (Ciocca, Napoletano, & Schettini, 2018; He, Zhang, Ren, & Sun, 2016).

Food recognition flourished with the improvement of computational performance and the computer vision and machine learning advances over the last decade. Matsuda et al., in 2012, achieved accuracy scores of 55.8% for multiple-item food images and 68.9% for single-item food images by utilizing two different methods (Matsuda, Hoashi, & Yanai, 2012). The first method was based on the Felzenszwalb's deformable part model and the second was based on feature fusion. In 2014 one of the first works that employed deep learning was developed (Kawano & Yanai, 2014b). The researchers achieved a classification accuracy of 72.26% with a pre-trained (transfer learning) model similar to AlexNet (Krizhevsky, Sutskever, & Hinton, 2012). At the same period, Bossard et al. achieved 50.76% classification accuracy with the use of random forest on the Food101 dataset (Bossard, Guillaumin, & Van Gool, 2014). The researchers noted that the random forest method cannot outperform deep learning approaches. Similar conclusions were drawn by Kagaya et al., who reported a classification accuracy of 73.70% using deep Convolutional Neural Networks (CNNs) (Kagaya, Aizawa, & Ogawa, 2014). The next year, deep CNNs, in combination with transfer learning, achieved 78.77% (Yanai & Kawano, 2015). Christodoulidis et al. introduced a new method on deep CNNs achieving an accuracy of 84.90% on a custom dataset (Christodoulidis, Anthimopoulos, & Mougiakakou, 2015). In 2016, Singla et al. adopted the GoogLeNet architecture (Szegedy, et al., 2015) and with a pre-trained model they achieved an accuracy of 83.60% (Singla, Yuan, & Ebrahimi, 2016). Lui et al., in 2016, developed DeepFood and achieved similar results by using optimized convolution techniques in a modified version of the Inception (Liu, et al., 2016). The researchers reported an accuracy of 76.30% on the UEC-Food100 dataset, an accuracy of 54.70% on the UEC-Food256 dataset and an accuracy of 77.40% on the Food-101 dataset. Hassannejad et al. scored an accuracy of 81.45% on the UEC-Food100 dataset, 76.17% on the UEC-Food256 dataset and 88.28% on the Food-101 dataset (Hassannejad, et al., 2016), using Google's image recognition architecture named Inception V3 (Szegedy, Vanhoucke, Ioffe, Shlens, & Wojna, 2016).

In 2017, Ciocca et al. introduced a novel method that combined segmentation and recognition techniques based on deep CNNs on a new dataset achieving 78.30% accuracy (Ciocca, Napoletano, & Schettini, 2017a). Mezgec et al. (2017) adopted a modification of the popular AlexNet and introduced the NutriNet, which was trained with images acquired using web search engines. They reached a classification performance of 86.72% over 520 food and drinks classes. The NutriNet uses fewer parameters compared to the original structure of the AlexNet. In 2018, Ciocca et al. introduced a new dataset, which was acquired from the merging of other datasets (Ciocca, Napoletano, & Schettini, 2018). They tested several popular architectures on the dataset, including a Residual Network (He, Zhang, Ren, & Sun, 2016) with 50 layers, setting the latter as the reference architecture. Apparently, many different research groups have worked vigorously on the topic of food recognition, using a variety of methods and techniques. Overall, the most successful methods are the variations of deep learning approaches.

This chapter serves the purpose of presenting the possible paths to follow in order to develop a machine learning technique for the complex task of food recognition. It starts with an introduction of general deep learning concepts to establish a basic level of understanding and an outline of the most popular deep learning frameworks available for the design and development of new models. Next, the three possible development approaches are being presented, including a totally new architecture, several transfer learning adaptations and the most relevant content prediction platforms. As the empirical learning employed in deep learning is data demanding, popular image datasets for food recognition are being presented in a separate subsection. A comparative study of the presented solutions follows, and the chapter concludes by providing information regarding the development of a deep learning model tasked to tackle the food recognition problem, as well as a list of future challenges in the relevant scientific area.

Kiourt, C., Pavlidis, G. and Markantonatou, S., (2020), **Deep learning approaches in food recognition**, *MACHINE LEARNING PARADIGMS - Advances in Theory and Applications of Deep Learning*, Springer

## 2. Background

The most recent food recognition systems are developed based on deep convolutional neural network architectures (Ciocca, Napoletano, & Schettini, 2017b; Ciocca, Napoletano, & Schettini, CNN-based features for retrieval and classification of food images, 2018; Horiguchi, Amano, Ogawa, & Aizawa, 2018; Yu, Anzawa, Amano, Ogawa, & Aizawa, 2018; Weiqing, Shuqiang, Linhu, Yong, & Ramesh, 2019), thus, this section presents some "must-know" techniques and methods for the development and utilization of an efficient food recognition CNN model.

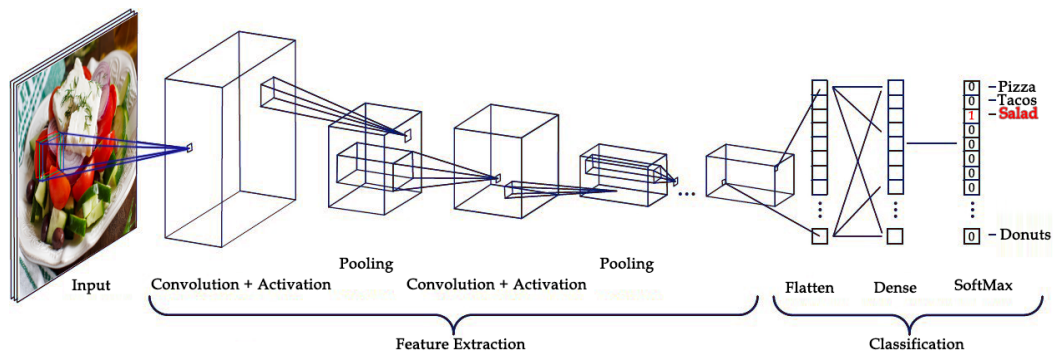

Figure 1. A simple CNN architecture

A CNN is a multi-layered neural network with a unique architecture designed to extract increasingly complex features of the data at each layer, in order to correctly determine the output. CNN's are well suited for perceptual tasks. Figure 1 depicts a simple CNN architecture with two convolution layers and two pooling layers at the feature extraction level of the network. The classification level of the network (top layers) is composed of three layers, (a) a flattening layer, which transforms the multiple feature maps, namely a multi-dimensional array (tensor) of the last convolution layer to a one-dimension array, (b) a dense layer and (c) a prediction layer. The output of the prediction layer is, in most cases, *one-hot encoded*, which means that the output is a binary vector representation of the categorical classes. The cell in this vector which is triggered (is true or "1") shows the prediction of the CNN, thus, each class is represented by a vector of "0"s and a single "1", the position of which determines the class.

A CNN is mostly used when there is an unstructured dataset (e.g., images) and the model needs to extract information from it. For instance, if the task is to predict an image caption:

- the CNN receives a color image (i.e. a salad) as a 3D matrix of pixels (three matrices of 2D intensity images);
- during the training, the hidden layers enable the CNN to identify unique features by shaping suitable filters that extract information from the images;
- when the network learning process converges, then the CNN it is able to provide a prediction regarding the class an image belongs to.

Each convolution in a network consist of, at least, an *input*, a *kernel* (filter) and a *feature map* (output).

- The *input* (or the input feature map) for a CNN is typically a tensor[1]. A tensor is a multidimensional array containing data for the validation of training of the CNN. An input, as a tensor, can be represented by $(i, h, w, c)$, where: $i$ is the number of images, $h$ is the height of the image, $w$ is the width of the image and $c$ is the number of channels in the image. For example, $(64, 229, 229, 3)$ represents a batch of 64 images of $229 \times 299$ pixels and 3 color channels (RGB).

- A *kernel* (also called filter and feature detector) is a small array with fixed values. Convolution of this array with the input tensor reveals visual features (such as edges). The kernel moves with specific steps and direction in order to cover the entire surface of each image. These step patterns are called the *strides*. A stride denotes the number of pixels by which the kernel moves after each operation.

- The *output feature map* or simply the output of a convolution is the group of features extracted by the filter (kernel) from the input, and is practically a transformed form of the input feature map. The size of the output feature map's tensor depends mainly on the input, the kernel size and the stride.

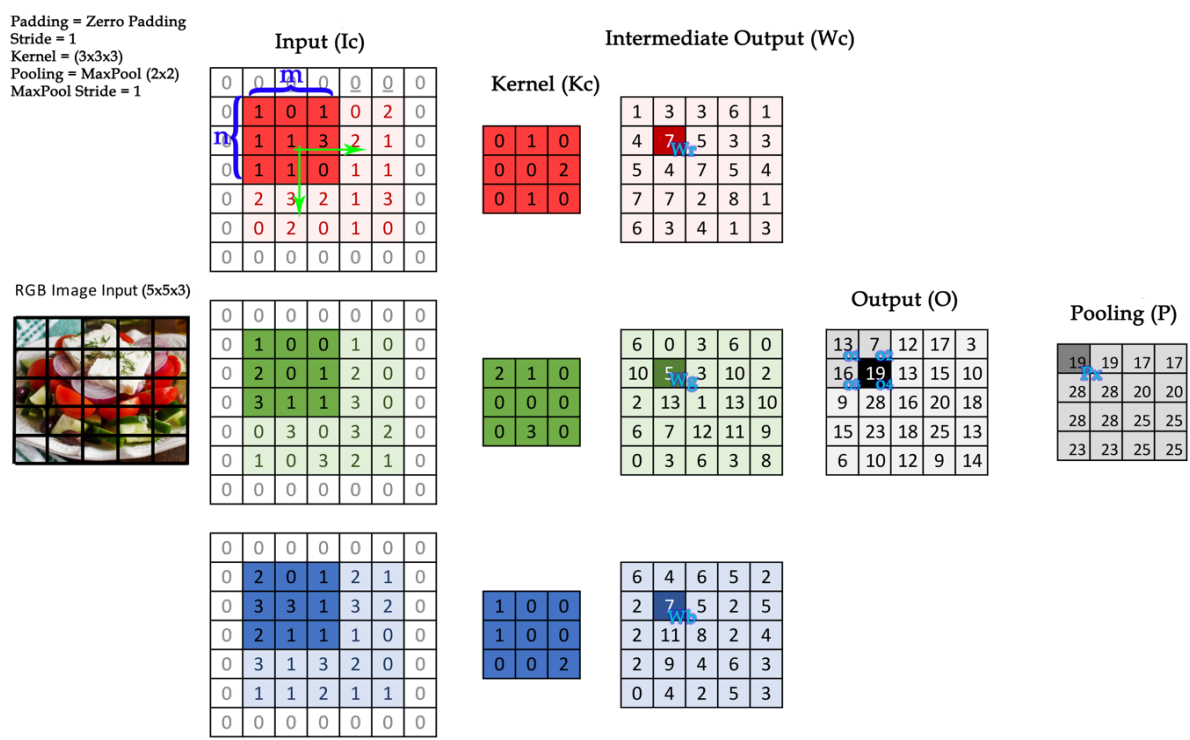

Figure 2. An analysis of the convolution process

---

[1] It should be noted that there is an inconsistency across the scientific domains in the usage of the word "tensor", which has a totally different meaning in physics than in computer science and in particular in machine learning.

Figure 2 depicts the entire process (a single step) for an RGB image, over $5 \times 5 \times 3$ image blocks. For each color channel $c$ of the input tensor an intermediate output ($\mathbf{W}_c$) is calculated as

$$\mathbf{W}_c = \mathbf{K}_c \otimes \mathbf{I}_{c\,(m,n)} \tag{1}$$

where $\otimes$ denotes the convolution, $\mathbf{K}_c$ is the convolution kernel and $\mathbf{I}_{c\,(m,n)}$ is the input image block.

The final feature map $\mathbf{O}$ is calculated by adding the intermediate outputs ($\mathbf{W}_c$) of each channel:

$$\mathbf{O} = \sum_c \mathbf{W}_c \tag{2}$$

In order to reduce the size of the feature maps, a pooling method is commonly used. Pooling reduces the size of a feature map by using some functions to summarize subregions. Typical functions used for the task are the maximum (max pooling) or the average (average pooling) of a subregion (or submap) $x$ of the feature map. In the example of Figure 2 max pooling is used, which is expressed as:

$$\mathbf{O}_x = \begin{bmatrix} o_{1,1} & o_{1,2} \\ o_{2,1} & o_{2,2} \end{bmatrix}, \qquad \mathbf{P}_x = \max_{i,j=1,2} o_x^{i,j} \tag{3}$$

where $o_x^{i,j}$ denotes the $i$-th, $j$-th element of the submap $\mathbf{O}_x$. Generally, the pooling window (subregion of the feature map) moves with a specific stride.

Most of the times, the size of a kernel is not consistent with the size of the input tensor, so to maintain the dimensions of the output (feature map) and in order for the kernel to be able to pass correctly over the entire input tensor, a data padding strategy is required. This is typical in any image convolution application. Padding is a process of adding extra values around the input tensor. One of the most frequent padding methods is *zero-padding*, which adds $z$ zeroes to each side of the boundaries of the input tensor. There are many different ways to calculate the number $z$. A simple way, if a stride of 1 is used, is to compute the size of zero padding as $z = \left\lceil \frac{(k-1)}{2} \right\rceil$, where $k$ is the filter size and $\lceil * \rceil$ the ceiling operation. For example, a kernel of size $3 \times 3$ imposes a zero-padding in each dimension of $z = \frac{(3-1)}{2} = 1$ zero, which adds a zero at the beginning and a zero at the end of each image dimension. Based on these values the size of the output of a convolution can be calculated easily. Supposing a one-dimensional case or a square input image and filter kernels, the formula to calculate the output size of any convolutional layer is given as

$$o = \left\lfloor \frac{i - k + 2z}{s} \right\rfloor + 1 \tag{5}$$

where $o$ is the dimension of the output, $i$ is the input dimension, $k$ is the filter size, $z$ is the padding, $s$ is the stride and $\lfloor * \rfloor$ the floor operation. For example, in a typical case in which $i = 5$, $k = 3$, $z = \lceil \frac{k-1}{2} \rceil = 1$ and $s = 2$, then the expected output should be of size $o = \lfloor \frac{5-3+2}{2} \rfloor + 1 = 2 + 1 = 3$. This example is graphically depicted in Figure 3, which shows the first two steps in the process of a square image convolution with a square filter kernel with the appropriate zero-padding.

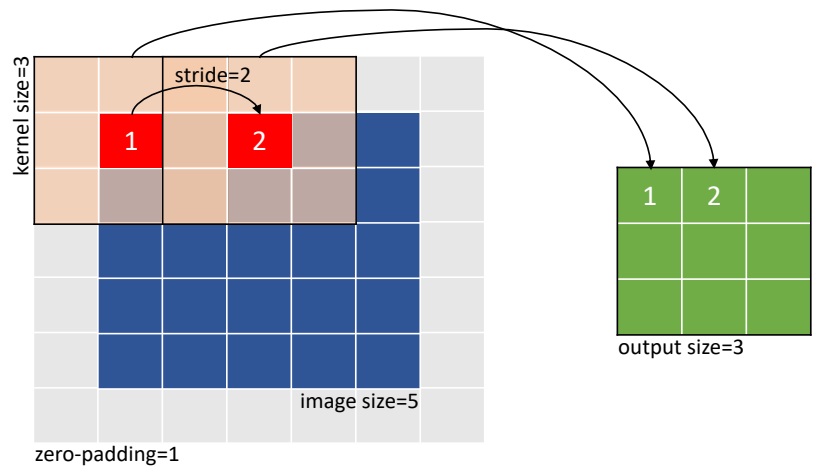

Figure 3. Example of a square image convolution with zero-padding

While training a CNN there are many parameters that should be taken into consideration for an effective training process, which will eventually produce a powerful and efficient deep learning model. Especially when a model has to handle a couple of hundreds to a million of trainable parameters, there are many challenges to be met. Among the most prominent issues are *overfitting* and *underfitting* (Goodfellow, Bengio, & Courville, 2016). There are a few ways to mitigate these issues. Underfitting occurs when the model fails to sufficiently learn the problem and performs poorly both on the training and the validation samples (lacks adaptation). Overfitting occurs when the model adapts too well to the training data but does not perform well on the validation samples (lacks generalization). Apparently, the ideal situation is when the model performs well on both training and validation data. So, it can be said that when a model learns suitably the training samples of the dataset and generalizes well, and if the model performs well in the validation samples, then there is high probability the process resulted in a *good-fitting model*. Simplistic examples of these issues are depicted in the graphs of Figure 4. The leftmost graph represents the dataset, including training and validation samples. Second from the left is an example of underfitting, whereas third from the left is an example of overfitting. The rightmost graph represents a good-fitting model. It should be noted that there are many ways to calculate the *error* in these issues, which in the graphs of Figure 4 are shown based on the validation data as red vertical lines.

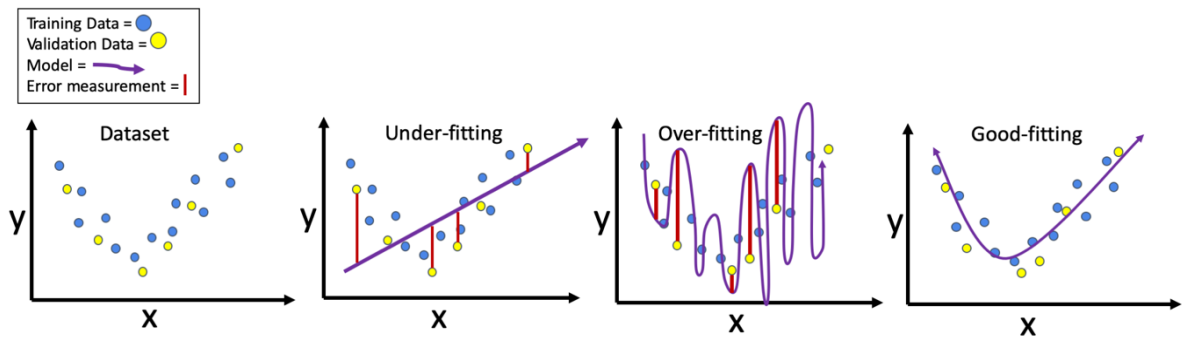

Figure 4. Underfitting, overfitting and good fitting examples

Some very common approaches to address the underfitting issue are to increase the dataset and train for a longer time, to check for a more proper regularization and to check if the model is not powerful enough (the architecture needs to change). On the other hand, using popular CNN architectures, which contain many convolutional layers with potentially hundreds to millions of trainable parameters, the probability of overfitting is very high. A solution for this problem is to increase the flexibility of the model, which may be achieved with regularization techniques, such the L1/L2 regularization (Andrew, 2004), the Dropout method (Srivastava, Hinton, Krizhevsky, Sutskever, & Salakhutdinov, 2014) and Batch Normalization (Ioffe & Szegedy, 2015).

The *L1 regularization*, uses a penalty technique to minimize the sum of the absolute values of the parameters, while the *L2 regularization* minimizes the sum of the squares of the parameters. Both L1 and L2 regularizations are known as "weight decay" methods because they try to keep the weights of the network as small as possible. The *Dropout* (Srivastava, Hinton, Krizhevsky, Sutskever, & Salakhutdinov, 2014), randomly removes neurons during training (units are randomly dropped, along with their connections), so the network does not over-rely on any particular neuron. *Batch Normalization* (Ioffe & Szegedy, 2015) is mainly used to increase the training speed and the performance of the network, but it can also be used to perform a regularization similar to dropout. Batch normalization is used to normalize the inputs of each layer (intermediate layers), in order to tackle the *internal covariate shift*[2] problem. It also allows the use of higher learning rates and, many times, acts as a powerful regularizer, in some cases eliminating the need for Dropout.

Apart from the methods presented in the previous paragraphs, data augmentation is another popular technique that effectively addresses both underfitting or overfitting issues. Deep learning models usually need a vast amount of data to be properly trained. So, the data augmentation technique is a very important part of the training process. Some of the most popular operations that can be applied in images to perform data augmentation include (a) flipping, (b) random cropping, (c) tilting, (d) color shifting,

---

[2] Covariate shift relates to a change in the distribution of the input variables of the training and the test data.

(e) rotation, (f) corruption with noise, and (g) changing of contrast. It should be noted that when the number of classes is large and the network is too complex, a combination of these methods may be the appropriate approach to obtain a good-fitting model.

For an in-depth treatment of the convolutional neural networks and the apprehension of the mathematical background the interested reader is advised to consider (Zeiler & Fergus, 2014; Dumoulin & Visin, 2016; Goodfellow, Bengio, & Courville, 2016; Chollet, Deep Learning with Python, 2018), or the available vast relevant literature and online sources.

## 2.1 Popular deep learning frameworks

A deep learning framework is a platform containing interfaces, libraries, tools and applications, which allows developers and scientists to easily build and deploy complex deep learning models, without directly being involved in designing their architectures and handling the math. Also, deep learning frameworks offer building blocks for designing, training and validating deep neural network models, through high level programming interfaces. Most of these frameworks rely on GPU-accelerated solutions such as CuDNN to deliver high-performance multi-GPU accelerated training. Some key features of a good deep learning framework are:

- optimization for performance
- GPU- acceleration
- good documentation and good community support
- easy to code
- parallelized processes
- ready to use deep models/architectures (applications)
- pre-trained models
- multiple, ready to use functions (e.g. optimizers, activation functions, layers etc.)
- popularity
- continues update

Based on these key features six deep learning frameworks stand out and are presented in this section. Figure 5 presents a simple hierarchical view of these frameworks, highlighting how all of them are GPU accelerated. It should be stressed that only the Keras framework can be considered as high-level API developed over other libraries, mainly focusing on the provision of a user-friendly and easy to extend environment (framework).

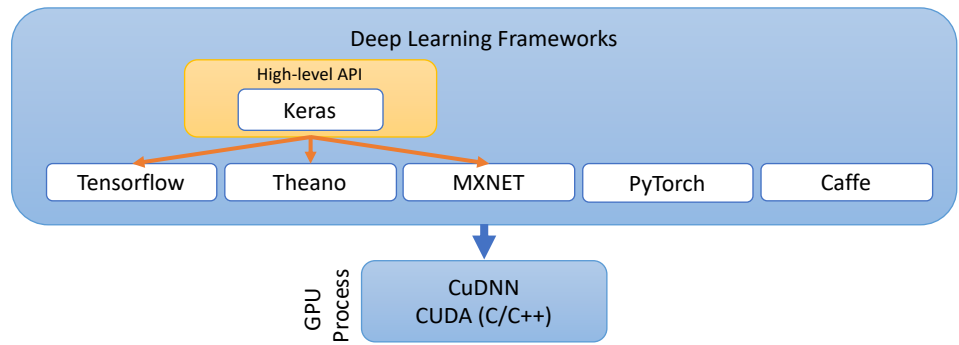

Figure 5. Most popular deep learning frameworks

*Caffe*[3] (Jia, et al., 2014) is the oldest and the most popular deep learning framework developed by Yangqing Jia at BAIR (Berkeley Artificial Intelligence Research). Caffe supports a layer-by-layer deep neural network development. While a new layer could be developed in Python programming language it is considered to be less effective in contrast to a layer written in C++ CUDA. Caffe besides the Python API provides also a MATLAB interface. Several years after its first release, its godfather (Yangqing Jia) and his team at Facebook developed the next generation Caffe framework in 2018, called Caffe2. This version was improved towards the development of a more user-friendly framework. In addition, it provides an API that supports mobile applications. Also, Caffe2 supports the Open Neural Network Exchange (ONNX) format, which allows easy integration of models with other frameworks, such as MXNet, PyTorch etc. In fact, since May 02, 2018, Caffe2 was integrated in PyTorch.

*TensorFlow*[4] (Abadi, et al., 2016) is an open source framework developed and backed by a machine learning team in Google. Its development and design are based on numerical computations using data flow graphs focusing on large-scale distributed training models. It consists of more than two hundred standard operations, including a variety of mathematical operations, multi-dimensional array manipulation, flow control and many more, written in C++ programming language. This framework is GPU-accelerated and may also run in CPUs and mobile devices. For this reason, it could be considered as a framework for use both in scientific research and in product development. Its distinctive drawback is that it requires a low-level API and its use seems quite difficult to inexperienced users. However, it has been integrated into several high-level APIs, such as Keras, Estimator and Eager.

*Keras*[5] (Chollet, François Chollet, 2015) is an open source neural network library running on top of some popular frameworks such as TensorFlow and Theano. It is considered to be a high-level API (interface), written in Python, rather than a standalone machine-learning framework, focusing on its extensibility, its modularity and its user-friendly environment. It requires minimum knowledge and technical skills. Since 2017, the TensorFlow team includes Keras in TensorFlow's core library.

---

[3] http://caffe.berkeleyvision.org/
[4] https://www.tensorflow.org/
[5] https://keras.io/

*PyTorch*[6] (Adam Paszke, 2017) is a GPU-accelerated deep learning framework written in a combination of programming languages, Python, C and CUDA. It was developed by the Facebook AI research team. PyTorch is younger than TensorFlow but it has grown rapidly in popularity, because of its flexibility and the ease of development when it comes to complex architectures. Additionally, it allows customizations that TensorFlow does not. The framework is used by both scientific and industrial communities, for example Uber has been using PyTorch for its applications. PyTorch is backed by some of the most high-tech companies such as Facebook, Twitter and NVIDIA. It should be highlighted that PyTorch supports ONNX models, allowing an easy exchange of models among a variety of frameworks.

*Theano*[7] (TheanoTeam, 2016) was first released in 2007 and it has been developed at the University of Montreal. It is the second oldest significant Python deep learning framework. Theano compiles mathematical expressions by using NumPy in Python and some efficient native libraries such as BLAS to run the experiments in multiple GPUs and CPUs. In November 2017, Theano stopped to be actively upgrading, but it is maintained continuously. Note, that Keras runs also on top of Theano.

*MXNet*[8] (Chen, et al., 2015) is incubated by Apache and used by Amazon. It is the fifth most popular deep learning library. One important benefit of MXNet is its portability, its lightweight development, and its scalability, which allows for an effective adaptation in a variety of computer systems with multiple GPUs and CPUs. This key point makes MXNet very powerful and useful for enterprises. Also, its portability supports a wide range of smart devices, such as mobile devices (using Amalgamation), IoT devices (supporting AWS Greengrass) and Serverless (using AWS Lambda). Furthermore, it is cloud-friendly and directly compatible with a variety of cloud services. MXNet supports Keras as a top layer interface, which allows easy and fast prototyping. MXNet supports ONNX format models.

There are several comprehensive studies (Jovic, K., & Bogunovic, 2014; Nguyen, et al., 2019) presenting the benefits and drawbacks of various frameworks/libraries focused on machine learning. It should be mentioned that most frameworks have a scientifically documented presence. A brief presentation of the six aforementioned frameworks is provided in the following paragraphs. All of them mainly focus on deep learning neural networks and are shorted based on their popularity according to the literature in Table 1. Observe that, while Theano is the oldest framework, it is not the most popular. The most popular framework in the scientific community is Caffe. Moreover, Python programming interface is common for all the frameworks. An interested reader is advised to consider the extensive survey published in 2019 by Nguyen et al. that considers almost all machine learning frameworks and libraries (Nguyen, et al., 2019).

---

[6] https://pytorch.org/
[7] http://www.deeplearning.net/software/theano/
[8] https://mxnet.apache.org/

Table 1. Most popular deep learning frameworks

| Name | Released | Programming Interfaces | Citations[9] |
|---|---|---|---|
| Caffe | 2014 | Python, MATLAB, C++ | 11990 |
| TensorFlow | 2016 | Python, C++[a], Java[a], Go[a] | 9250 |
| Keras | 2015 | Python | 5399 |
| PyTorch | 2017 | Python, ONNX, C++ | 2718 |
| MXNET | 2015 | Python, C++, R, Java, ..[b] | 1077 |
| Theano | 2007 | Python, C++ | 572 |

[a] Not fully covered
[b] Many more languages

## 3. Deep learning methods for food recognition

Nowadays, practitioners of deep learning debate on choosing to develop new models or to adopt existing ones and sometimes, opinions are sharply divided. In summary, the development of a new deep learning model (architecture) applied to a new domain (problem - task) is a complex and time-consuming process that requires programming and mathematical skills beyond the average, along an understanding of data-driven (empirical) model learning. On the other hand, transfer learning with fine-tuning should be adopted when an immediate solution is required. This approach is less demanding in technical skills and theoretical knowledge that could be basically limited to the development of a simple classification method. In addition, there are today a number of online platforms that enable the rapid development of new models based on proprietary technologies, which basically act as black boxes to the developers. These platforms provide a third alternative and are typically machine learning systems pre-trained with large datasets spanning a wide spectrum of different classes ready to provide solutions to a wide variety of problems, through the use of Application Programming Interfaces (APIs). This section reviews these three approaches in the context of food recognition in mobile tourism applications. The study begins with the presentation of the relevant image datasets.

### 3.1 Food image datasets

Deep learning approaches, especially CNN methods, require large datasets to build an efficient classification model. Nowadays, even if there are hundreds of datasets available, with a variety of content, datasets focusing on nutritional content (foods, drinks, ingredients etc.) are very limited and quite small (Ciocca, Napoletano, & Schettini, 2017b). In this section we present some of the most popular and large datasets focusing in food recognition used in the literature. Table 2 summarizes these datasets and ranks them by the number of the images they contain and the year they have been introduced. It should be noted, that the last five years numerus teams have worked on the development of larger food datasets, since the recognition of foods and their ingredients is a very interesting and

---

[9] Google Scholar: Last accessed 7 Nov 2019

challenging problem with many real-life applications (Mezgec & Koroušić, 2017) (Ciocca, Napoletano, & Schettini, 2017b) (Ciocca, Napoletano, & Schettini, 2017a) (Kuang-Huei, Xiaodong, Lei, & Linjun, 2018)

Table 2. Most popular large datasets focusing on food categories

| Name | #Classes | #Images | Year | Description |
|---|---|---|---|---|
| UECFood100 | 100 | 14,461 | 2012 | Most categories are popular foods in Japan |
| UECFood256 | 256 | 31,651 | 2014 | |
| Food-101 | 101 | 101,000 | 2015 | More general food categories |
| UMPCFood-101 | 101 | 100,000 | 2015 | |
| VireoFood-172 | 172 | 110,241 | 2016 | |
| Food524DB | 524 | 247,636 | 2017 | |
| Food-101N | 101 | 310,000 | 2018 | |

Deep Learning approaches mainly provide a prediction for each sample (image) based on mathematical probabilities. In most cases, the evaluation of a prediction is based on the comparison of the network result (prediction) and the true class label. The *top-1 accuracy* is a sorting of the results of a prediction that reveals if the target output (class) is the same with the top class (the one with the highest calculated probability) of the prediction made for an image by the model. *Top-k accuracy* (Zhang, et al., 2017) can be defined as

$$\text{top-k} = \frac{1}{N}\sum_{i=1}^{N} 1[a_i \in C_i^k] \qquad (6)$$

where $1[\cdot] \rightarrow \{0,1\}$ denotes the indicator function. If the condition in the square brackets is satisfied, which stands for $a_i$ as the ground truth answer to a question $q_i$, and $C_i^k$ as the candidate set with top-k of the highest similar answers, then the function returns 1, or 0 otherwise.

The *UECFood100*[10] dataset (Matsuda, Hoashi, & Yanai, 2012) was first introduced in 2012 and contains 100 classes of food images. In each image the food item is located within a bounding box. Most of the food categories (classes) are popular Japanese foods. For this reason, most of the food classes might not be familiar to everyone. The main development reason of this dataset was to implement a practical food recognition system to use in Japan. In 2014 Kawano and Yanai used this dataset to score a classification rate of 59.6% for the top-1 accuracy and 82.9% for the top-5 accuracy, respectively (Kawano & Yanai, 2014). The same team improved their classification results, achieving 72.26% in the top-1 accuracy and 92,00% in the top-5 accuracy when using a Deep CNN pre-trained with the ILSVRC2010[11] dataset (Kawano & Yanai, 2014b).

---

[10] http://foodcam.mobi/dataset100.html
[11] http://www.image-net.org/challenges/LSVRC/2010/

The *UECFood256*[12] dataset (Kawano & Yanai, 2014a) consists of 256 classes of food images. The structure of the classes and the images arew similar to the *UECFood100*. Again, most food categories are focused on Japanese foods. This dataset was used in order to develop a mobile application for food recognition. In 2018 Martinel, Foresti and Micheloni achieved a classification performance of 83.15% for the top-1accuracy and 95.45% for the top-5 accuracy on this dataset (Martinel, Foresti, & Micheloni, 2018).

The *Food-101*[13] dataset (Bossard, Guillaumin, & Van Gool, 2014) was introduced in 2014 from the Computer Vision Lab, ETH Zurich, Switzerland. In total, the dataset has a thousand images for each of the 101 food classes (categories), comprising a dataset of 101,100 images. For each class, 250 test images and 750 training images are provided. All images were manually reviewed and training images contain some noise on purpose, such as intense image color shift and wrong labels. All images were rescaled to have a maximum side length of 512 pixels. In 2018 Martinel, Foresti and Micheloni reported 90.27% for the top-1 and 98.71% for the top-5 on the food-101 dataset (Martinel, Foresti, & Micheloni, 2018).

The *UPMC Food-101* (Wang, D. Kumar, Thome, Cord, & Precioso, 2015) was developed in 2015 and is considered as a large multimodal dataset consisted of 100,000 food items classified in 101 categories. This dataset was collected from the web and can be considered as a "twin dataset" to the Food-101, since they share the same 101 classes and they have approximately the same size.

*VireoFood-172*[14] (Chen & NGO, 2016) is a dataset with images depicting food types and ingredients. This data set consist of 110,241 food images from 172 categories. All images are manually annotated according to 353 ingredients. The creators of this dataset used a MultiTaskDCNN and achieved classification rates of 82.05% for the top-1 accuracy and 95.88% for the top-5 accuracy (Chen & NGO, 2016).

The *Food524DB*[15] (Ciocca, Napoletano, & Schettini, 2017b) was introduced 2017 and is considered to be one of the largest publicly available food datasets, with 524 food classes and 247,636 images. This dataset was developed by merging food classes from existing datasets. The dataset has been constructed by merging four benchmark datasets: VireoFood-172, Food-101, Food50, and a modified version of UECFood256. The creators of this dataset used the popular RestNet-50 architecture and scored an 81.34% for the top-1 accuracy and an 95.45% for the top-5 accuracy (Ciocca, Napoletano, & Schettini, 2017b). A modification of this dataset was released in 2018 as Food-475, introducing small differences (Ciocca, Napoletano, & Schettini, 2018).

---

[12] http://foodcam.mobi/dataset256.html
[13] https://www.vision.ee.ethz.ch/datasets_extra/food-101/
[14] http://vireo.cs.cityu.edu.hk/VireoFood172/
[15] http://www.ivl.disco.unimib.it/activities/food524db/

*Food-101N*[16] (Kuang-Huei, Xiaodong, Lei, & Linjun, 2018) was introduced in 2018 in CVPR from Microsoft AI & Research. The dataset was designed to address label noise with minimum human supervision. This dataset contains 310,000 images of food recipes classified in 101 classes (categories). Food-101and Food-101N and UPMC Food-101 datasets, share the same 101 classes. Additionally, the images of Food-101N are much noisier. Kuang-Huei, Xiaodong and Linjun provided an image classification of 83.95% for the top-1 accuracy by using the CleanNet model (Kuang-Huei, Xiaodong, Lei, & Linjun, 2018).

Note that the development of a new large datasets with world-wide known food categories, is a very difficult and time-consuming process. Nowadays, multiple research teams focus on automated processes for the development of larger datasets. Automated processes employ powerful search image engines, crowdsourcing and social media content to perform the task of data collection. For example, Mezgec et al. (2017) used an image search engine to develop a food dataset and the result was quite interesting.

## 3.2 Approach #1: new architecture development

This section presents a CNN model (named *PureFoodNet*) that was developed for the purposes of an R&D project in Greece that targets regional Greek food recognition for mobile tourism applications. This model is based on the building blocks and principles (simple homogenous topology) of the VGG architecture (Simonyan & Zisserman, 2015). A common practice for developing new CNN architectures (models) is to add convolutional layers, until the model starts performing well and overfits. After that, the tuning process is applied on the hyperparameters of the model until a good-fitting, high accuracy model emerges. This process is time-consuming and needs vast computational resources. For example, the new model presented, the PureFoodNet, required a full month's time to be developed and properly configured on a high-end computing system.

The architecture of the PureFoodNet is depicted in Figure 6. It consists of three convolutional blocks and a classification block. The first block contains two convolutional layers each having 128 filters. The second block contains three convolutional layers with 256 filters each and the third block contains three convolutional layers with 512 filters each. The last block which is usually called the "Top Layers" or the "Classification layers" contains a dense layer with 512 neurons and a predictor layer (Dense) with 101 outputs. For the initial development of this model, note that, the output of the prediction layer was designed to match the 101 classes (food categories) of the Food101 dataset. In order to avoid overfitting in the PureFoodNet model, the architecture includes classical regularization mechanisms such as Dropout, L2 regularization and Batch Normalization layers. Also, for the purpose of the PureFoodNet's tuning, a weight initialization method was used (Glorot & Bengio, 2010; Sutskever, Martens, Dahl, & Hinton, 2013). To opt for a fast and accurate training, an optimizer with Nestorov momentum was employed (Sutskever, Martens, Dahl, & Hinton, 2013).

---

[16] https://kuanghuei.github.io/Food-101N

In food recognition and in many other cases, it is important to know what each layer (or even each neuron) learns. One method to do this is by visualizing the layers' values (local "knowledge"). Hence, the issue of the unused (dead) filters can be addressed easily. This issue can be noticed by checking if some activation maps are all zero for many different inputs, which is a symptom caused by a high learning rate. The most commonly used method is to visualize the activations of the layers during the forward pass. Networks that use ReLU activation functions (Glorot, Bordes, & Bengio, 2011) usually have activations that look too noisy (relatively blobby and dense) in the beginning of the training. After several training epochs and after the network starts to perform well, the visualization of the activations becomes more intuitive, sparse and localized. In Figure 6, above each block, a few activations of some indicative layers are shown in order to highlight the information learnt in each layer. It is clear that the lower one looks (Block 1) at the network the more general information the layer keeps, in contrast to the high-level layers (Block 3) where more specific/targeted information is learnt by the network. This is because in many images of food dishes a self-similarity exists in the form of repeating patterns of the same shapes throughout the surface of the image. This is why it is expected that the tiling of an image caused by the convolutional layers improves the localization of the ingredients, which, in turn, results a better learning of the dishes and improved predictions.

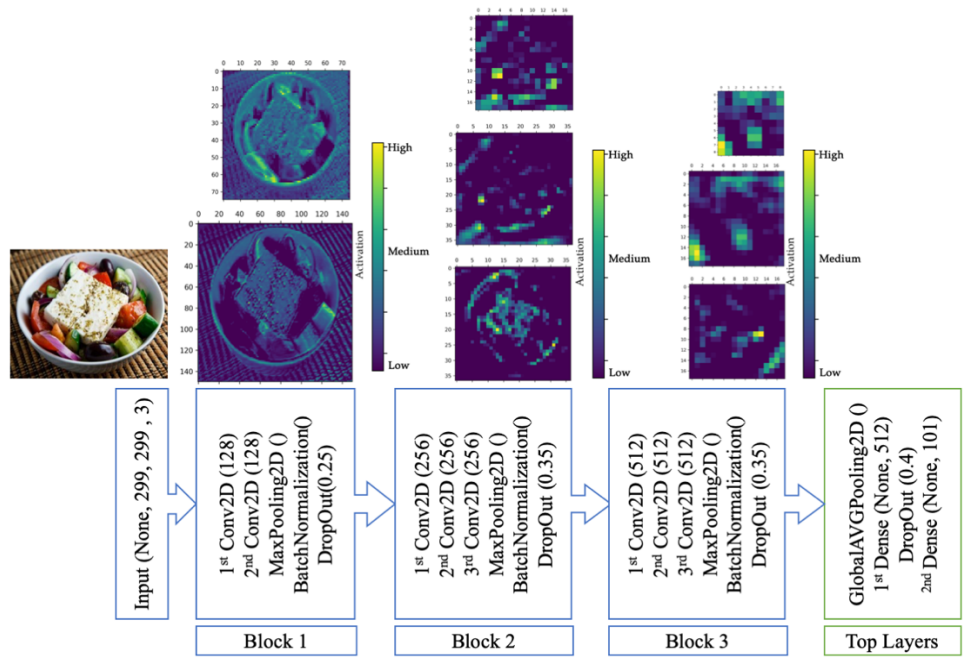

Figure 6. PureFoodNet: a simple pure CNN architecture for food recognition

## 3.3 Approach #2: transfer learning and fine-tuning

The second alternative path in deep learning-based food recognition approaches, relies on using popular pre-trained deep learning models, implementing the transfer learning technique. Transfer learning in deep learning theory is a popular machine learning technique, in which a model developed and trained for a specific task is reused for another similar task under some parametrization (fine tuning). An intuitive definition of transfer learning was provided by Torrey and Shavlink in (2009): "*Transfer learning is the improvement of learning in a new task through the transfer of knowledge from a related task that has already been learned*". Fine tuning is the process in which a pre-trained model is used without its top layers, which are replaced by new layers, more appropriate to the new task (dataset). For example, if the pre-trained PureFoodNet model were to be used, the Top Layer should be removed (i.e. the layers inside the green rectangular in Figure 6) and only the convolutional layers should be kept with their weights. New top layers should be added according to the needs of the new task (new dataset) and the number of the outputs in the prediction layer should be associated with the number of the new classes. It is a common practice, to make an appropriate recombination of the top layers, a reconfiguration of the parameters, and many repetitions in order to achieve a desirable performance.

Because of the ease of use of transfer learning, nowadays, very few people develop and train new CNN models (architectures) from scratch. This is due to the fact that it is difficult to have a dataset of sufficient size for the specific task. Also, as mentioned before, the development of a new model is complex and time-consuming. Instead, it is more convenient to reuse a model that was trained on a very large dataset, like the ImageNet (Zeiler, 2013) that contains more than 14 million images spanning more than 20,000 categories altogether. Also note that, the ImageNet contains a lot of categories with images depicting foods or edible items in general, thus a pre-trained model on the ImageNet dataset is expected to be quite effective and easily tuned to improve the performance in food recognition applications.

## 3.4 Approach #3: deep learning platforms

The need for the development of a new model or the fine-tuning of an existing pre-trained model for a specific task is sometimes questionable, due to the existence of a third alternative in deep learning practice: the powerful online platforms, which are pre-trained with large datasets on a wide spectrum of classes. In this section some popular platforms for image-based food recognition are presented, along with an assessment on their efficacy and a summary of their benefits and drawbacks. These platforms keep on training on a daily basis and thus are improving over time. Some of these platforms belong to well-established technology-providing enterprises, who benefit from the deep learning services and applications. These enterprises offer the option for commercial use of their infrastructures to the scientific and industrial communities either in the form of a paid subscription or under limited free trial usage, which is, in most cases, quite enough for experimentation. It is emphasized, though, that the pre-trained models offered by these platforms work as "black boxes", since they do not provide enough details about their prediction mechanism, architectures etc. and their parameterization is restricted. All these platforms provide several tools and programming interfaces through APIs.

The *Vision AI*[17] (VAI) was developed by Google as a cloud service in 2016. It includes multiple different features detection, such as Optical Character Recognition (OCR) boosted with a mechanism focusing on handwritten text recognition in multiple languages, face detection, moderate content (explicit content such as adult, violent, etc.), landmarks and logo detections.

The *Clarifai*[18] (CAI) was developed in 2013 for the ImageNet Large Scale Visual Recognition Competition (ILSVRC) and won the top 5 places that year (Zeiler, 2013). Clarifai is the only platform that provides options regarding the model selection for the prediction. It offers quite an interesting range of models applied for different purposes, such as generic image recognition, face identification, Not Safe for Watching (NSFW) recognition, color detection, celebrity recognition, food recognition, etc.

*Amazon Rekognition*[19] (AR) was first introduced in 2016 by Amazon as a service after the acquisition of Orbeus, a deep learning startup. Amazon Rekognition provides identification of objects, people, text, scenes, and activities, and it is able to detect any inappropriate content, with considerable capabilities in facial detection and analysis (emotion, age range, eyes open, glasses, facial hair, etc.).

*Computer Vision*[20] (CV) was developed by Microsoft as part of the Cognitive Services. This platform offers detection of color, face, emotions, celebrities, text and NSFW content in images. An important feature of Computer Vision is the OCR system, which identifies text in 25 different languages, along with a powerful content interpretation system.

## 4. Comparative study

This section presents a comparative study for the three aforementioned alternatives in developing deep learning models for image-based food recognition. The introduced architecture of the PureFoodNet is tested against the most popular pre-trained models on the Food101 dataset, whereas the deep learning platforms are tested against each other in a toy experiment that reveals their strengths and weaknesses.

### 4.1 New architecture against pre-trained models

Before anything, the PureFoodNet has a much shallower architecture than the rest of the pre-trained models against which it is compared. In the development of the PureFoodNet random weight initialization was used, in order to speed up the training process, and a custom adaptation of the learning rate. For the experiments presented in this section VGG16 (Simonyan & Zisserman, 2015), InceptionV3 (Szegedy, Vanhoucke, Ioffe, Shlens, & Wojna, 2016), ResNet50 (He, Zhang, Ren, & Sun, 2016), InceptionResNetV2 (Szegedy, Ioffe, Vanhoucke, & Alemi, 2017), MobileNetV2 (Sandler, Howard, Zhu, Zhmoginov, & C.L., 2018), DenseNet121 (Huang, Liu, van der Maaten, & Weinberger, 2017), and NASNetLarge (Zoph, Vasudevan, Shlens, & Le, 2018) deep CNN architectures are being considered. The python code for the experiments can be accessed through a source code management repository[21].

---

[17] https://cloud.google.com/vision/
[18] https://clarifai.com
[19] https://aws.amazon.com/rekognition/
[20] https://www.microsoft.com/cognitive-services

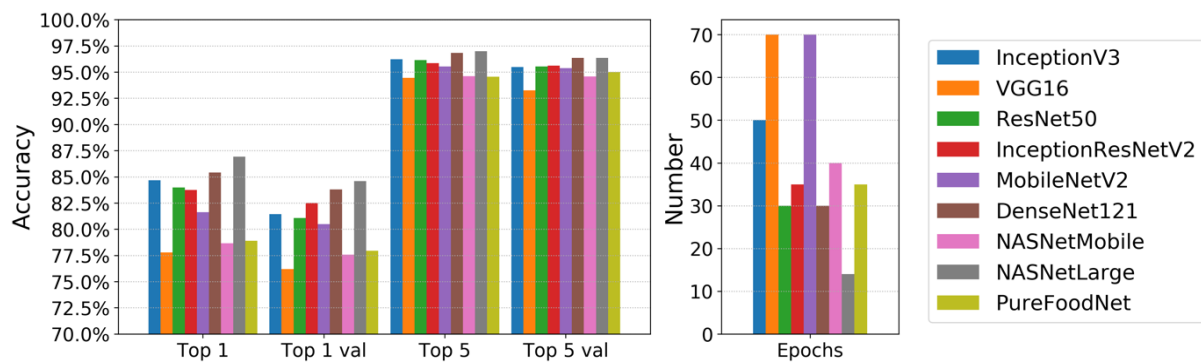

Figure 7. Overview of the comparison of popular deep architectures against PureFoodNet in food recognition

In these experiments, each model used exactly the same top layers (the layers in the green rectangle in Figure 6), in order to have a fair comparison. Figure 7 presents a bar graph of the accuracy of each model, as well as a bar graph of the number of epochs needed for the models to achieve these results. The accuracy bar graph depicts the categorical top-1 and top-5 performances, both for the training (top-1, top-5) and for the validation (top-1 val, top-5 val) of each model. Training of the models, in these experiments, halted the moment the validation accuracy stopped improving, thus the difference in the required training epochs reported for the models. It is evident that the majority of the models performed quite well. The worst validation accuracy observed was 76% for the VGG16 model, which was also among the slowest to learn. An interesting observation, which is not apparent in the graphs, is related to the performance of the NASNetLarge that achieved the best score with a very few training epochs, however, requiring approximately 83,33% more time than any other pre-trained network used in this experiment and around 92% more time than the PureFoodNet. Clearly PureFoodNet performed well even though it was not the best. There are cases in which the difference in performance between PureFoodNet and NASNetLarge was insignificant, especially from the perspective of top-5 accuracy.

Depending on the application, one should always consider the complexity of each model, thus there is a trade-off between performance and requirements/specifications, and a balance should be reached. The model size produced by NASNetLarge is about 86% larger than the size of PureFoodNet. In addition, in a study that examines the task of food recognition, the analysis that each layer provides should be in line with the semantic meaning in the nature of food's structure, namely the ingredients. Due to this, the efficiency of a model in this task is not precisely quantified by typical measures and metrics. Consequently, the choice of a model (architecture) cannot solely rely on typical performance scores (accuracy achieved during the training process). An elegant model selection is a complicated procedure, which should take into account more criteria, besides class prediction.

---

[21] https://github.com/chairiq/FoodCNNs

## 4.2 Deep learning platforms against each other

In the following paragraphs a toy experiment is unfolded to provide intuition on the evaluation of popular deep learning platforms in image-based food recognition. Two different test images (Figure 8) with complicated content were selected for the experiment. Both images contain multiple ingredients. Image A (Figure 8 left) presents a Greek salad. This image contains some noticeable ingredients, like the sliced tomatoes, the cucumber, the onion, the cheese (feta), the olives, the dill and the parsley. Image B (Figure 8 right) presents a fruit dessert, in which there are strawberries, blueberries, kiwis, grapefruits, oranges and a cream at the bottom of the cup.

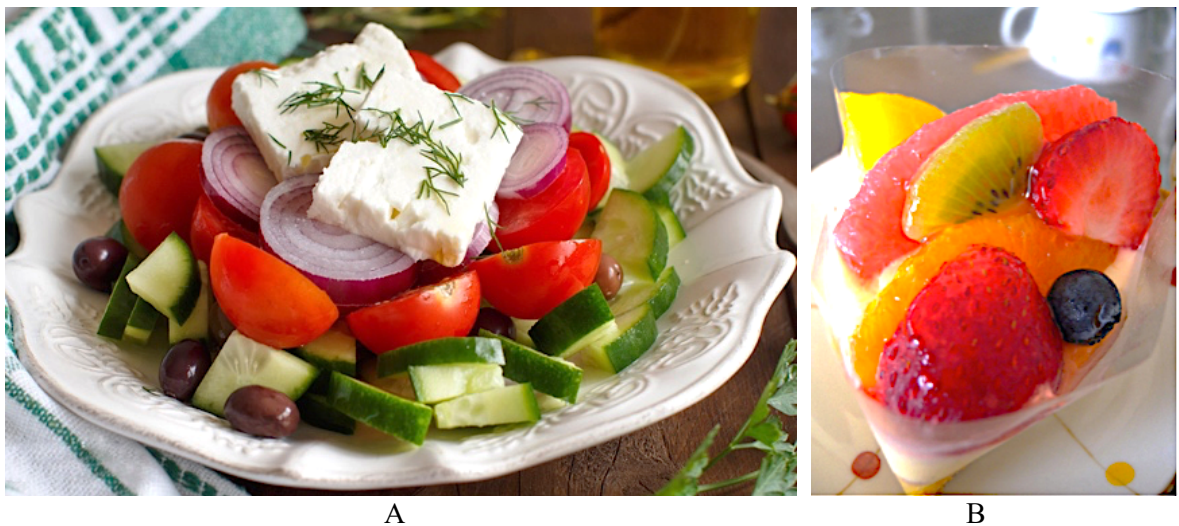

A                                                                B
Figure 8. Test images: (A) Greek salad and (B) Fruit dessert

Figure 9 and Figure 10 present the prediction results for the two test images respectively by all reviewed platforms, with (a) green bars corresponding to true positive (correctly predicted) classes (in terms of ingredients, food type, etc.), (b) red bars corresponding to false positive (wrongly predicted) classes and (c) blue bars corresponding to neutral, partially correct or unimportant (to this case study) results.

Even a superficial comparison of the results presented in Figure 9, clearly shows that both CV and AR perform poor, both in term of the ingredients and in terms of the overall results. VAI was somewhat better but still identified some non-existing (or irrelevant to the context) classes. CAI performed significantly better but also reported non-existing food ingredients within the image. Apparently, the number of the correctly identified classes in combination with their accuracy, clearly rank CAI as the best platform for image A. It should be noted that CAI was the only platform that recognized the existence of parsley, which was out of the plate (right bottom in image A). VAI predicted with quite high accuracy that the food type might be a "Caprese Salad" (64% accuracy) or Israeli Salad (63% accuracy), and this is clearly justifiable because of the high similarity of these three salads.

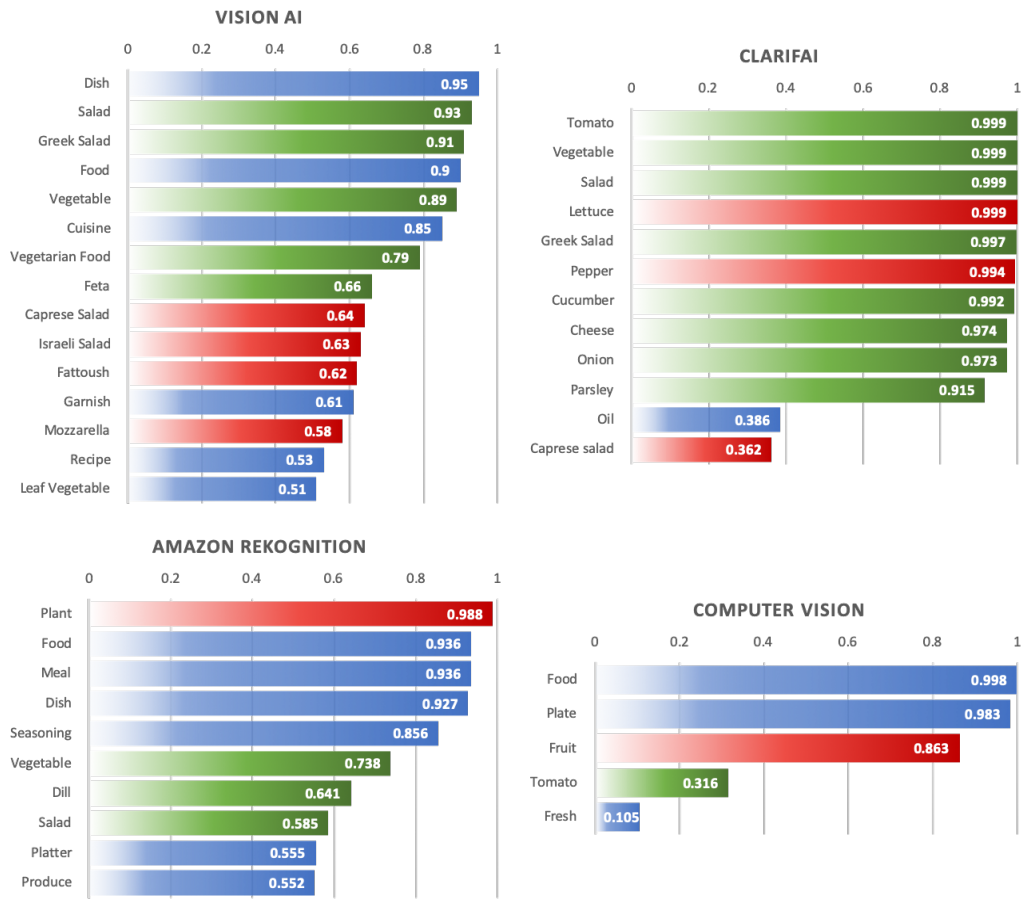

Figure 9. Results on food recognition image A by the reviewed platforms

In Figure 10, in which the results on image B are shown, it is once again clear that CV identified the smallest number of classes, including two very general classes (Fresh, Food). However, it is also important that CV is the only platform that spotted the class "Sliced" with very little confidence, though. AR did not perform well also, recognizing only a few correct classes and with a very low confidence. It is clear that CAI and VAI performed well in this case. Although CAI presented some false positive (but still somehow relevant) results the high confidence in true positives and the more classes reported render it a valuable tool for the task. On the other hand, VAI seems to have managed avoid false positives, but the number of classes and the confidence are considerably lower.

Overall, from just a toy experiment, CAI and VAI seem to perform better among the tested content identification platforms in the domain of food recognition, with slightly better results provided by CAI, apparently due to its specialized model on food. What is important from the developer perspective is that none of these platforms requires advanced programming skills to develop an image content analysis system.

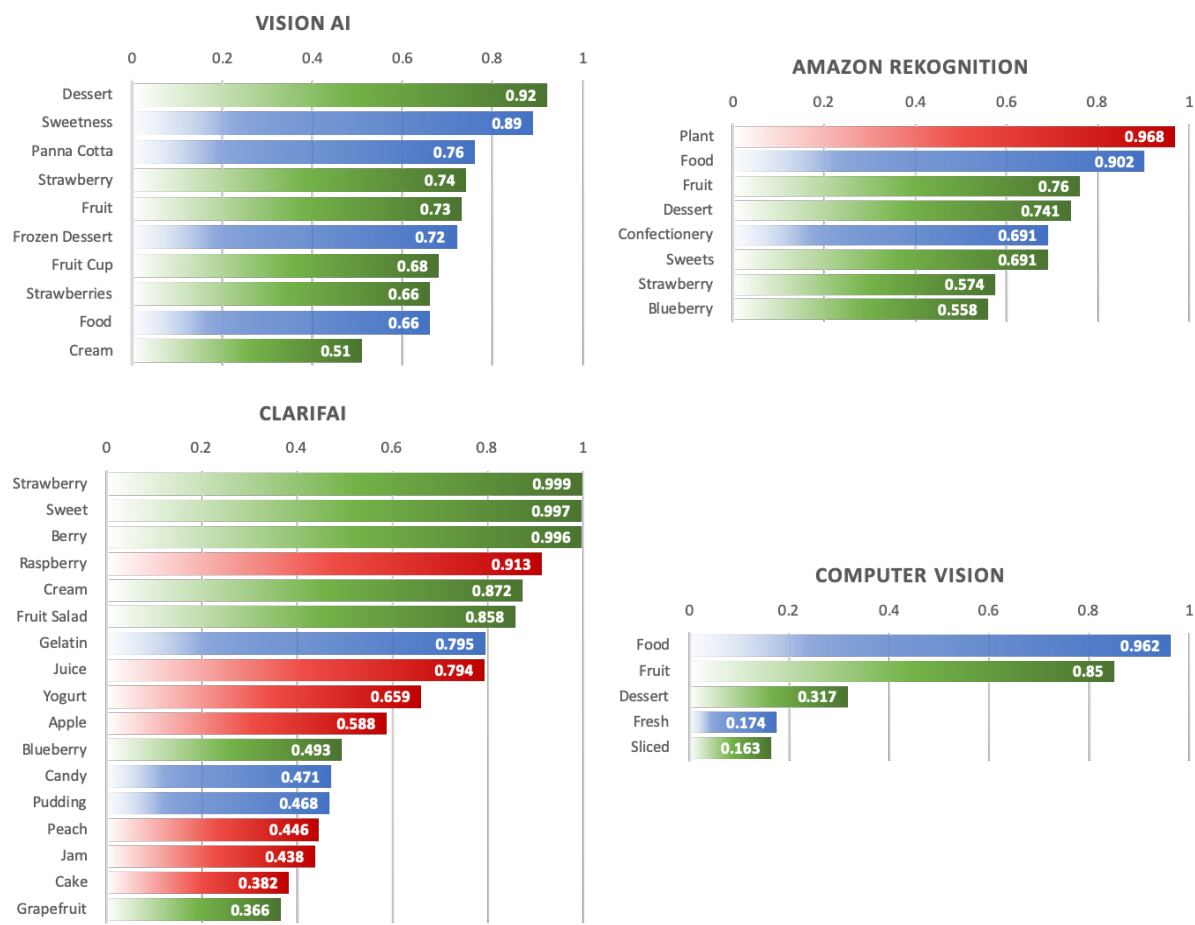

Figure 10. Results on food recognition image B by the reviewed platforms

# 5. Conclusions

Food recognition based on images is a challenging task that ignited the interest of the scientific communities of computer vision and artificial intelligence in the past decade. It is certain that a dish of food may contain a variety of ingredients, which may differ in shapes and sizes depending on the regional traditions and habits of the local communities. For example, it is worldwide known today that a salad with tomatoes, olives and onions relates to a Mediterranean diet. Thus, the distinction among a Greek, a Caprese or an Israeli salad might be a difficult task, even for humans. The way ingredients are sliced and placed on a plate may be a lead for a distinction, but this can also be debated.

In any modern machine learning task, it is often debatable whether to design a new deep learning model (architecture), or to utilize an existing model through transfer learning, or to use an available image-based content prediction platform (typically through APIs). Designing a new architecture requires extensive knowledge and intuition, whereas transfer learning, basically, requires specific knowledge and platform-based application development is just a technical solution. On the other hand, from the productivity and ease of deployment perspective (with a compromise in efficiency), the approaches are ranked in the opposite direction, with platforms easily ranking first.

This chapter focused on the particularly interesting and complex problem of image-based food recognition and tried to build on those three lines of machine learning solutions to reveal the strength and weaknesses of each approach and highlight the opportunities being offered and the inherent threats both in the approaches and the prerequisites. Furthermore, as machine learning approaches are data hungry, popular large datasets for food recognition were presented and the need for even better datasets with more specific and world-wide known food categories was identified. The study highlighted a need for the labeling of food ingredients, which would enable deep learning approaches to attain better performances and empower more useful real-world applications (like in nutrition and tourism). Overall, this chapter suggests a bright future in the domain of food recognition with deep learning approaches, and proposes a number of future directions such as:

- creation of large-scale food datasets, with specific content, world-wide known food data categories
- labeling of datasets not only with food types (classes), but also with food ingredients (multi labelled classes)
- development of hybrid methods for better results, such as combinations of new models with prediction platforms
- development and integration of food and ingredient ontologies
    - to assist more accurate predictions in general
    - to develop rules based on contextual factors that (a) may lead to the development of similar food separation rules based the origin, habits, culture, etc., (b) may assist in targeted predictions based on geographical locations, personal preferences, etc.

## Acknowledgements

This research has been co-financed by the European Regional Development Fund of the European Union and Greek national funds through the Operational Program Competitiveness, Entrepreneurship and Innovation, under the call RESEARCH – CREATE – INNOVATE (project code: T1EDK-02015). We also gratefully acknowledge the support of NVIDIA Corporation with the donation of the Titan Xp GPU used for the experiments of this research.

Kiourt, C., Pavlidis, G. and Markantonatou, S., (2020), **Deep learning approaches in food recognition**, *MACHINE LEARNING PARADIGMS - Advances in Theory and Applications of Deep Learning*, Springer